\pdfoutput=1

\documentclass[11pt]{article}

\usepackage[preprint]{acl}

\usepackage{times}
\usepackage{latexsym}
\usepackage{amsmath}
\usepackage{tikz}
\usepackage{graphicx}
\usepackage{subcaption}
\usepackage{float} 
\usepackage{multirow}
\usepackage{booktabs}
\usepackage{url}

\usepackage[T1]{fontenc}

\usepackage[utf8]{inputenc}

\usepackage{microtype}

\usepackage{inconsolata}

\usepackage{graphicx}

\title{Is Bigger Edit Batch Size Always Better? - An Empirical Study on \\Model Editing with Llama-3}

\author{Junsang Yoon, Akshat Gupta, Gopala Anumanchipalli \\
UC Berkeley \\
  \texttt{\{junyoon, akshat.gupta, gopala\}@berkeley.edu}}

\begin{document}
\maketitle
\begin{abstract}
This study presents a targeted model editing analysis focused on the latest large language model, Llama-3. We explore the efficacy of popular model editing techniques - ROME, MEMIT, and EMMET, which are designed for precise layer interventions. We identify the most effective layers for targeted edits through an evaluation that encompasses up to 4096 edits across three distinct strategies: sequential editing, batch editing, and a hybrid approach we call as sequential-batch editing. Our findings indicate that increasing edit batch-sizes may degrade model performance more significantly than using smaller edit batches sequentially for equal number of edits. With this, we argue that sequential model editing is an important component for scaling model editing methods and future research should focus on methods that combine both batched and sequential editing. This observation suggests a potential limitation in current model editing methods which push towards bigger edit batch sizes, and we hope it paves way for future investigations into optimizing batch sizes and model editing performance.

\end{abstract}

\section{Introduction}

In the rapidly evolving field of artificial intelligence, keeping large language models (LLMs) up-to-date with the latest information presents a pivotal challenge. Traditional approaches often require retraining models on extensive datasets, a process that is both time-consuming and resource-intensive. An alternative is model editing \cite{editing-survey}, which allows for the modification of stored facts within a model, as well as the correction of inaccuracies. Several popular methods have emerged that infuse knowledge into models without the need for an additional hypernetwork, such as ROME (Rank-One Model Editing) \cite{ROME}, MEMIT (Mass Editing Memory in Transformer) \cite{MEMIT}, and EMMET (Equality-constrained Mass Model Editing algorithm for Transformers) \cite{akshat-unified}. These methods, traditionally called "locate-and-edit" algorithms, were recently shown to optimize the same objective, known as the \textbf{preservation-memorization} (PM) objective \cite{akshat-unified}. They directly modify specific "knowledge-containing" areas of the model without necessitating additional training, and are applicable to any transformer-based large language models (LLMs). In this work, we focus on parameter-modifying model-editing methods \cite{editing-survey} that do not require an additional hypernetwork \cite{hypernetwork}.

In this work, we present a step-by-step guide for using model editing methods based on the PM-objective for a new model. Since Llama-3 \cite{llama3} was recently released, we use it as an example to go through each decision point for model editing. First, we make edits on all Llama-3-8b layers to find the layer that gives best editing performance, creating a balance between editing accuracy and preserving existing knowledge. Once the optimal layer is found, we perform single layer editing experiments using ROME, MEMIT and EMMET. In our work, we make three different kinds of edits - singular edits, batched edits, as well as \textbf{sequential-batched} edits. A singular edit is when only one fact is edited inside a model at the time of evaluation. With batched edits, we update a \textit{batch-size} number of facts with a single update. In sequential-batched edits, we update batches of facts sequentially on the same model. This means the next batch of edits are made to the model containing the previous batch of edits. Prior work has focused on increasing editing capability by increasing the "batch size" \citep{MEMIT, MALMEN, akshat-unified} to scale model editing, but recent work has shown that this leads to severe model degradation \citep{hurt, akshat-catastrophic}. Thus, we ask the question - \textit{is increasing edit batch size the correct approach to scale model editing?} 

We compare the performance of batched model editing with sequential-batched editing. We find that for Llama-3, sequential-batched editing with a batch size of 1024 has optimal scaling performance, when compared to making simple batched-edits or sequential-batched edits with smaller batch size, thus showing that sequential model editing is an important component for large-scale model editing. Sequential model editing also enables model editing methods to approach the continual learning paradigm. With this study, we also provide baseline experiments on Llama-3 models to establish benchmarks for future research, as well as provide a transparent procedure for the different decision made while editing a model.\footnote{Our code is available here - \url{https://github.com/scalable-model-editing/unified-model-editing}}.




\section{Background}

\subsection{Preservation-memorization objective}\label{sec:pm}
\citet{akshat-unified} show that ROME and MEMIT both optimize the same objective function, called the \textbf{preservation-memorization} objective. The objective consists of two parts, a preservation term and a memorization term. The ROME optimization objective uses an equality-constrained for memorization as shown below:

\begin{equation}
\begin{aligned}
    & \underset{\hat{W}}{\operatorname{argmin}} \hspace{4pt} \underbrace{\left\| \hat{W} K_0 - W_0 K_0 \right\|}_{\text{preservation}} \hspace{4pt}  \text{    s.t. } \underbrace{\hat{W} k_e = v_e}_{\text{memorization}}
\end{aligned}
\end{equation}

Where $W$ represents the weights of the feed-forward layer we want to edit, $k$ is a key-vector representative of a fact, $v_e$ is the desired output, and $K_0 = [k^0_1 \hspace{4pt}| k^0_2 \hspace{4pt}| \dots |\hspace{4pt}k^0_N]$ is a matrix consisting of facts we want to preserve. The optimization leads to the ROME solution as follows: 

\begin{align}\label{eq:rome-update-equation}
    \hat{W} &= W_0 + \Delta \hspace{10pt} \text{where} \hspace{10pt}\\
    \Delta &= (v_e - W_0k_e) \frac{k_e^TC_0^{-1}}{k_e^TC_0^{-1}k_e}
\end{align}

MEMIT optimizes the same objectives but performance memorization using a least-squares constraint, which allows for a closed-form solution for making many memory edits with a single gradient updates, also known as batched-edits. The objective function for MEMIT is:

\begin{equation}
     \underset{\hat{W}}{\operatorname{argmin}} \hspace{4pt} \underbrace{\lambda|| \hat{W} K_0 - W_0 K_0 ||}_{\text{preservation}}  + \underbrace{||\hat{W} K_E - V_E ||}_{\text{memorization}}
\end{equation}

With $V_E$ again being a stacked matrix of $v_e$ vectors. In the above equations, a fact is represented by a pair of vectors called the \textit{key} ($k_e$) and \textit{value} ($v_e$) vectors. We refer the reader to prior works \citep{ROME, MEMIT, akshat-unified} for a more in-depth introduction of these methods. Again, this objective leads so a similar solution of the form: 

\begin{equation}\label{eq:memit}
\begin{aligned}
    \hat{W} &= W_0 + \Delta \hspace{10pt} \text{where} \hspace{10pt}  
    \\ \Delta &= \big(V_E - W_0K_E \big) K_E^T \big( \lambda C_0 + K_EK_E^T \big)^{-1}
\end{aligned}
\end{equation}

\citet{akshat-unified} also showed that it was possible to make batched edits using the equality constraint and present EMMET, an algorithm that allows for batched edits where memorization happens using an equality-constraint. The EMMET objective looks as follows:

\begin{equation}
\begin{aligned}
    \underset{\hat{W}}{\operatorname{argmin}} \hspace{4pt} \underbrace{\left\| \hat{W} K_0 - W_0 K_0 \right\|}_{\text{preservation}} \hspace{8pt}  \text{    s.t. } \\ \hspace{8pt} \underbrace{\hat{W} k^e_i = v^e_i \hspace{8pt} \forall i \in [1, 2 \dots E]}_{\text{memorization}}
\end{aligned}
\end{equation}

which, again, gives the solution:

\begin{equation}\label{eq:emmet_final}
\begin{aligned}
 &\hat{W} = W_0 + \Delta \hspace{10pt} \text{where} 
 \\ \Delta &= \left ( V_E - W_0 K_E \right ) \left ( K_E^T C^{-1}_0 K_E \right )^{-1} K_E^TC^{-1}_0 
 \end{aligned}
\end{equation}

\begin{figure*}[!ht]
    
    \centering
    \begin{subfigure}{.24\textwidth}
        \centering
        \includegraphics[width=\linewidth]{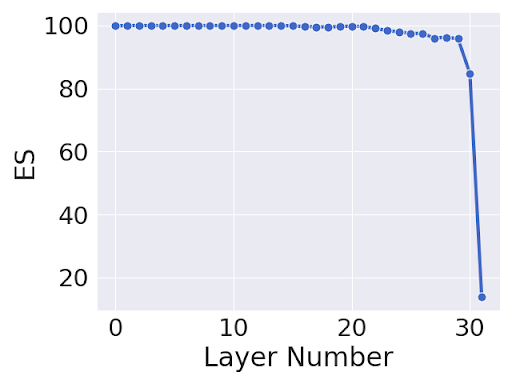}
        \caption{ES (MEMIT)}
        \label{fig:memit_gptj:edit_score}
    \end{subfigure}%
    \begin{subfigure}{.24\textwidth}
        \centering
        \includegraphics[width=\linewidth]{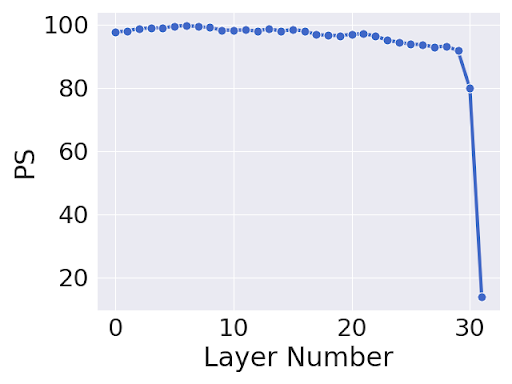}
        \caption{PS (MEMIT)}
        \label{fig:memit_gptj:downstream}
    \end{subfigure}%
    \begin{subfigure}{.24\textwidth}
        \centering
        \includegraphics[width=\linewidth]{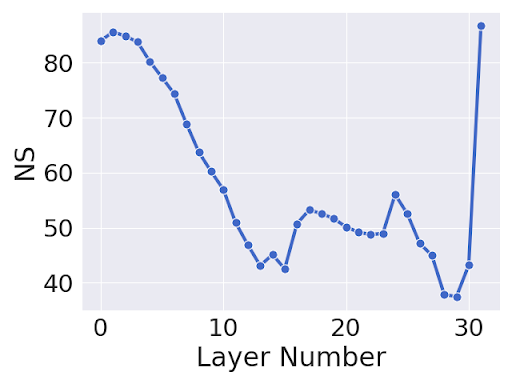}
        \caption{NS (MEMIT)}
        \label{fig:memit_gptj:downstream}
    \end{subfigure}%
    \begin{subfigure}{.24\textwidth}
        \centering
        \includegraphics[width=\linewidth]{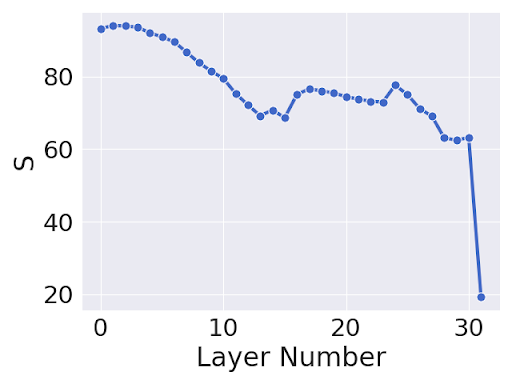}
        \caption{Score (MEMIT)}
        \label{fig:memit_gptj:downstream}
    \end{subfigure}

    \begin{subfigure}{.24\textwidth}
        \centering
        \includegraphics[width=\linewidth]{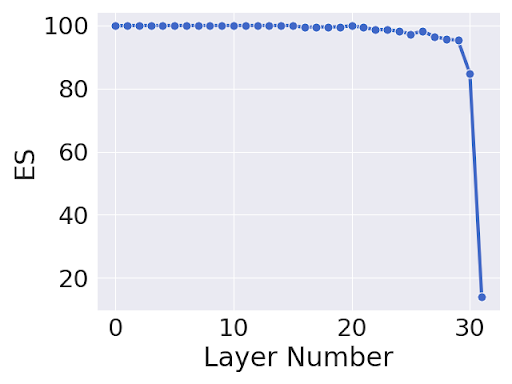}
        \caption{ES (ROME)}
        \label{fig:memit_gptj:edit_score}
    \end{subfigure}%
    \begin{subfigure}{.24\textwidth}
        \centering
        \includegraphics[width=\linewidth]{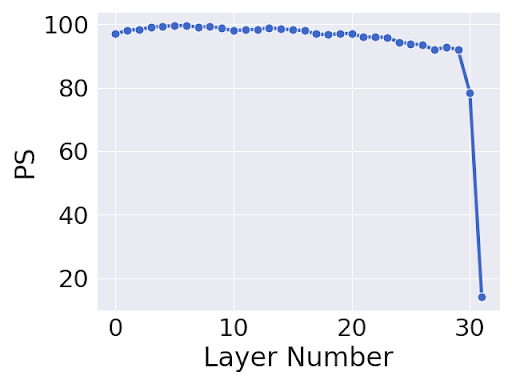}
        \caption{PS (ROME)}
        \label{fig:memit_gptj:downstream}
    \end{subfigure}%
    \begin{subfigure}{.24\textwidth}
        \centering
        \includegraphics[width=\linewidth]{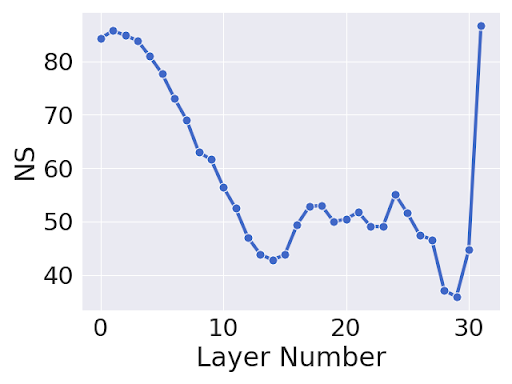}
        \caption{NS (ROME)}
        \label{fig:memit_gptj:downstream}
    \end{subfigure}%
    \begin{subfigure}{.24\textwidth}
        \centering
        \includegraphics[width=\linewidth]{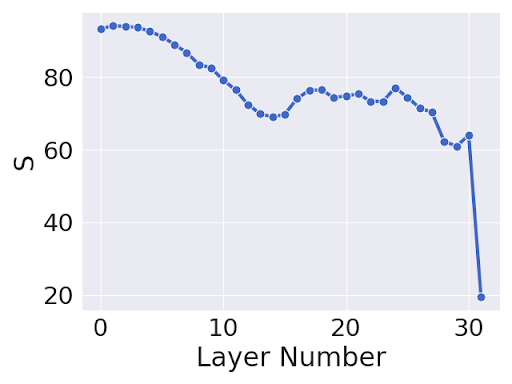}
        \caption{Score (ROME)}
        \label{fig:memit_gptj:downstream}
    \end{subfigure}

    \caption{Post-edit performance of various metrics for Llama3-8b model using MEMIT and ROME on various layers. To make the above plots, each layer is edited with 1000 facts, one at a time and non-sequentially.}
    \label{fig:layerwise-llama3}
\end{figure*}

\subsection{Model editing metrics}

Metrics to analyze the success of model edits are drawn from standard model editing metrics \cite{MEMIT, editing-survey}

\begin{itemize}

    \item \textbf{Efficacy Score (ES):} Measures the success of an edit within the model, measured by percentage where P(new fact) > P(old fact) for query prompt.

  \item \textbf{Paraphrase Score (PS):} A measure of a model's ability to generalize following an edit. Measured by where P(new fact) > P(old fact) under paraphrases of the query prompt.

  \item \textbf{Neighborhood Score (NS):} Represents the locality of model editing, measuring the impact of an edit on adjacent stored facts within the model. Specifically, NS quantifies the percentage of nearby facts that remain unchanged after an edit, thereby assessing the precision and isolation of the modifications.


\item \textbf{Composite Score (S):} Defined by \citet{ROME} as a holistic measure that combines aspects of edit success, generalization, and locality. It is calculated as the harmonic mean of the Edit Success (ES), Paraphrase Score (PS), and Neighborhood Score (NS), providing a comprehensive evaluation of the overall efficacy of model edits.

\end{itemize}

\section{Experiments}

\subsection{Finding Optimal Layer for Model Editing}
 \citet{MEMIT} assess the effectiveness of hidden states in LLMs for recalling facts using causal tracing \cite{vig2020causal}. They find that the representation of the subject's last token within the feed-forward networks (FFN) at intermediate layers plays a significant role. Building on this finding,  \citep{ROME, MEMIT} propose treating the linear layers as a key-value memory system, allowing for the modification of the values in effective hidden states to enhance memory recall. However, later work also showed that layers deemed important during causal tracing did not always translate to model editing performance \cite{doeslocalization}. Therefore, we find the optimal layer for model editing layer empirically.

\begin{table}
\centering

\scalebox{0.9}{
\begin{tabular}{@{}lccccc@{}}
\toprule
\multicolumn{6}{c}{Mean Scores for CounterFact Dataset} \\ \midrule

Algorithm & ES  & PS & NS  & GE & S  \\ \midrule
MEMIT     & 100.0 & 98.05 & 85.61 & 615.09 & 94.10 \\
ROME      & 100.0 & 98.05 & 85.73 & 614.42 & 94.15 \\
\bottomrule
\end{tabular}
}
\caption{Comparison of ROME and MEMIT on best-performing layer (layer 1) for singular edits on Llama-3. Here, at the time of evaluation, a model is edited with only 1 fact. The scores are calculated on a subset of 1k samples from the CounterFact dataset. }\label{tab:metrics_comparison}

\end{table}
 
 Specifically, we make 1000 non-sequential edits from the CounterFact \cite{ROME} dataset at each layer of the Llama-3 model. After implementing these edits, we calculate various model metrics to evaluate their impact. The overall score for each layer is derived from the harmonic mean of three key metrics: Efficacy Score (ES), Paraphrase Score (PS), and Neighbourhood Score (NS). The layer that achieves the highest score is selected as the most suitable for targeted interventions. Our findings, as shown in Figure \ref{fig:layerwise-llama3}, indicate that for Llama-3, layer 1 consistently outperforms on numerous metrics. Note here that Llama-3-8b layers are indexed from 0 to 31. This finding for Llama-3 is consistent with the previous version, Llama-2 \cite{llama2}, as seen in Figure \ref{fig:layerwise-llama2}. These results are contrary to the previous work \cite{editing-survey}, which suggest that layer 5 (0-indexed) is optimal for model editing for Llama-2. 

Figure \ref{fig:layerwise-llama3} also shows the both MEMIT and ROME have very similar performance for model editing across layers of a model. This resonates the fact that both algorithms optimize for the same objective with difference in the memorization constraint, and shows that this difference which has minor effect on editing performance. The least-square constraint allowed a closed-form solution for batched editing in MEMIT, which was also layer enabled with equality-constraint by \citet{akshat-unified} in the form of EMMET.

\begin{figure} 
    \centering
    \includegraphics[width=1.1\linewidth]{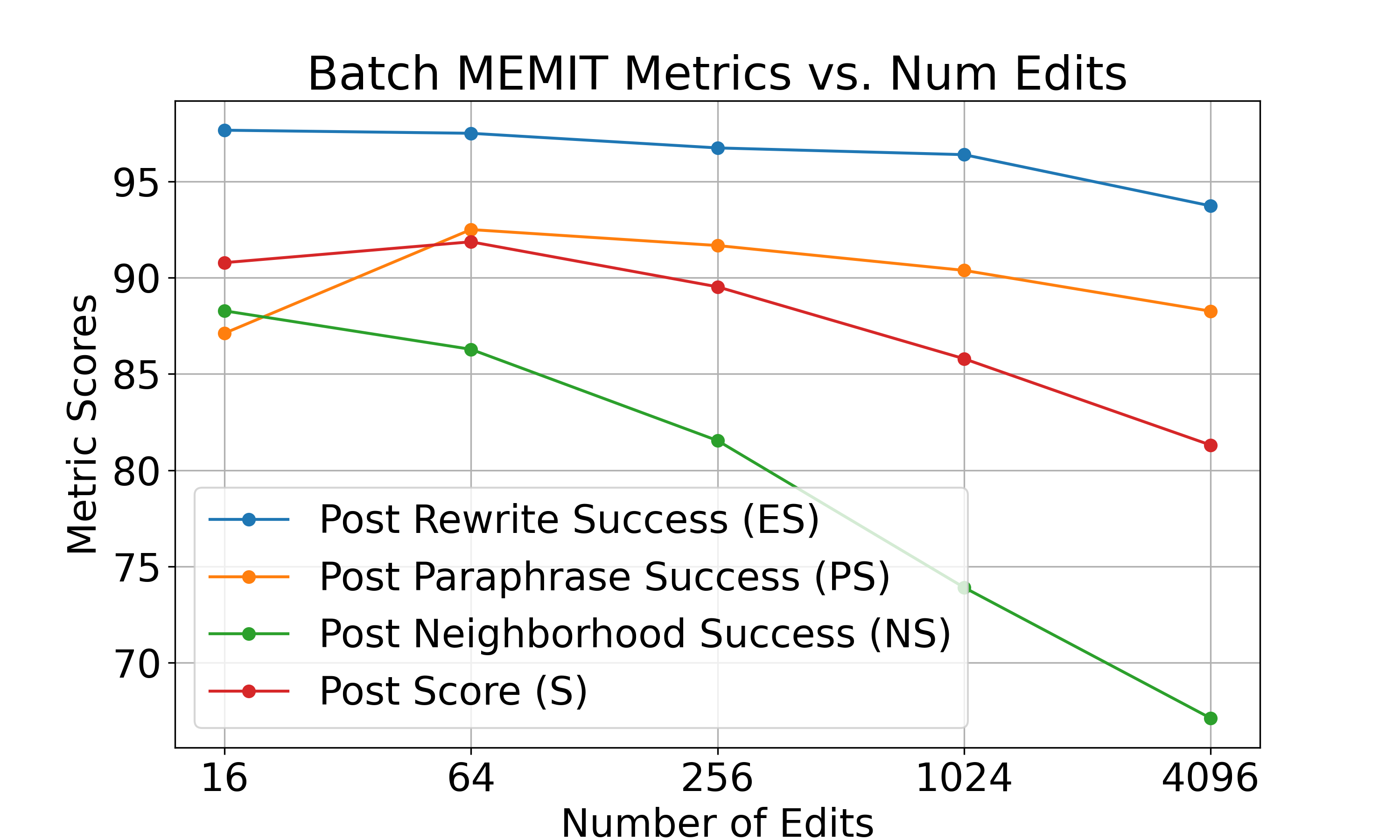}
    \caption{Figure shows various metric results (PS, NS, ES, S) after a batch MEMIT (16, 64, 256, 1024, 4096) edit.}
    \label{fig:batchMEMIT}
\end{figure}

\begin{figure} 
    \centering
    \includegraphics[width=1.1\linewidth]{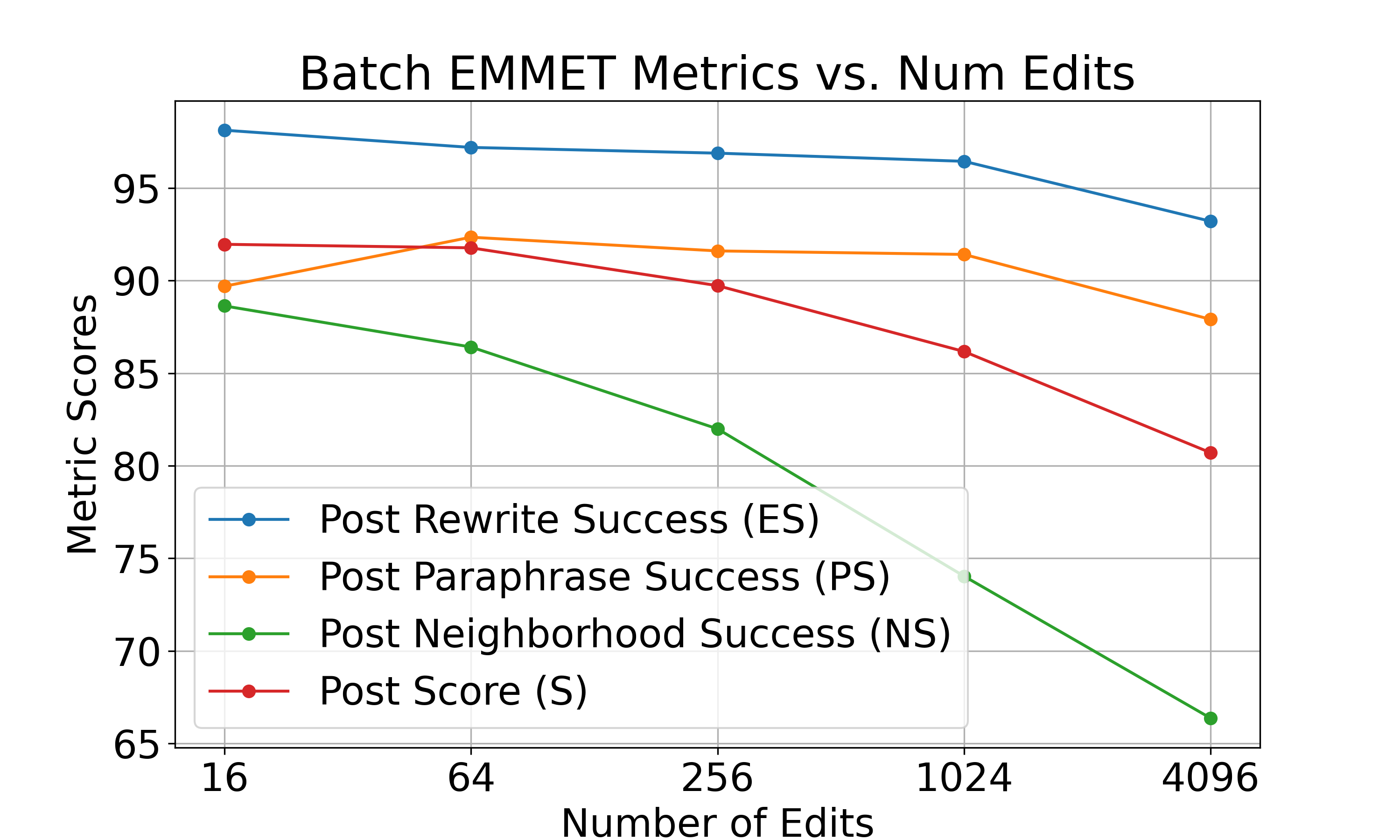}
    \caption{Figure shows various metric results (PS, NS, ES, S) after a batch EMMET (16, 64, 256, 1024, 4096) edit.}
    \label{fig:batchEMMET}
\end{figure}

\subsection{Batch Editing}
After finding the optimal layer for model editing, on to performing large scale model edits on the same model. One way of doing this is through batched editing. In batched edits, a large number of knowledge edits are performed on the model with the same update. \citet{akshat-unified} showed that editing multiple layers of a model can sometimes hide the efficacy of model editing performance, we stick to editing a single layer of the model. We edit layer 1 of Llama-3 with batch sizes of 16, 64, 256, 1024, and 4096 using MEMIT and EMMET. The hyperparameter tuning experiments for both algorithms can be found in \ref{sec:hparam-tuning}.

The evaluation results of Batch Editing with MEMIT is shown in Figure \ref{fig:batchMEMIT}. Metrics are seen to consistently fall with larger batches, with Neighbourhood Score (NS) being the most pronounced to fall. This suggests a heightened need to mitigate the impacts on locality following  model edits. Post Rewrite Success (ES) is shown to be the most resilient metric to edits. Post Paraphrase Success (PS) is actually first seen to increase dramatically between batch sizes of 16 and 64, the only metric to do so, suggesting a potential area for a possible investigation.

The evaluation results of Batch Editing with EMMET is shown in \ref{fig:batchMEMIT}. Similar to MEMIT. most metrics are seen to consistently fall with larger batches, with Neighbourhood Score again being the most pronounced to drop. Overall, the two methods show very similar trends, as reflected by the similarity in their optimization objectives.

\subsection{Sequential Batch Editing}
Above experiments showed the as batch size of edits increase, the model editing performance decreases significantly. This is especially true for the NS metric, showing that the edits made for larger batch sizes start to bleed into other facts known by the model. An alternate way to scale up model editing is sequential editing, where facts are added sequentially to a model. Thus, we ask the question - "Is there an optimal way to scale model editing that strikes a balance between these methods?"

\begin{figure*}[!ht]
    \centering
    
    \begin{subfigure}[b]{0.24\textwidth}
        \centering
        \includegraphics[width=\textwidth]{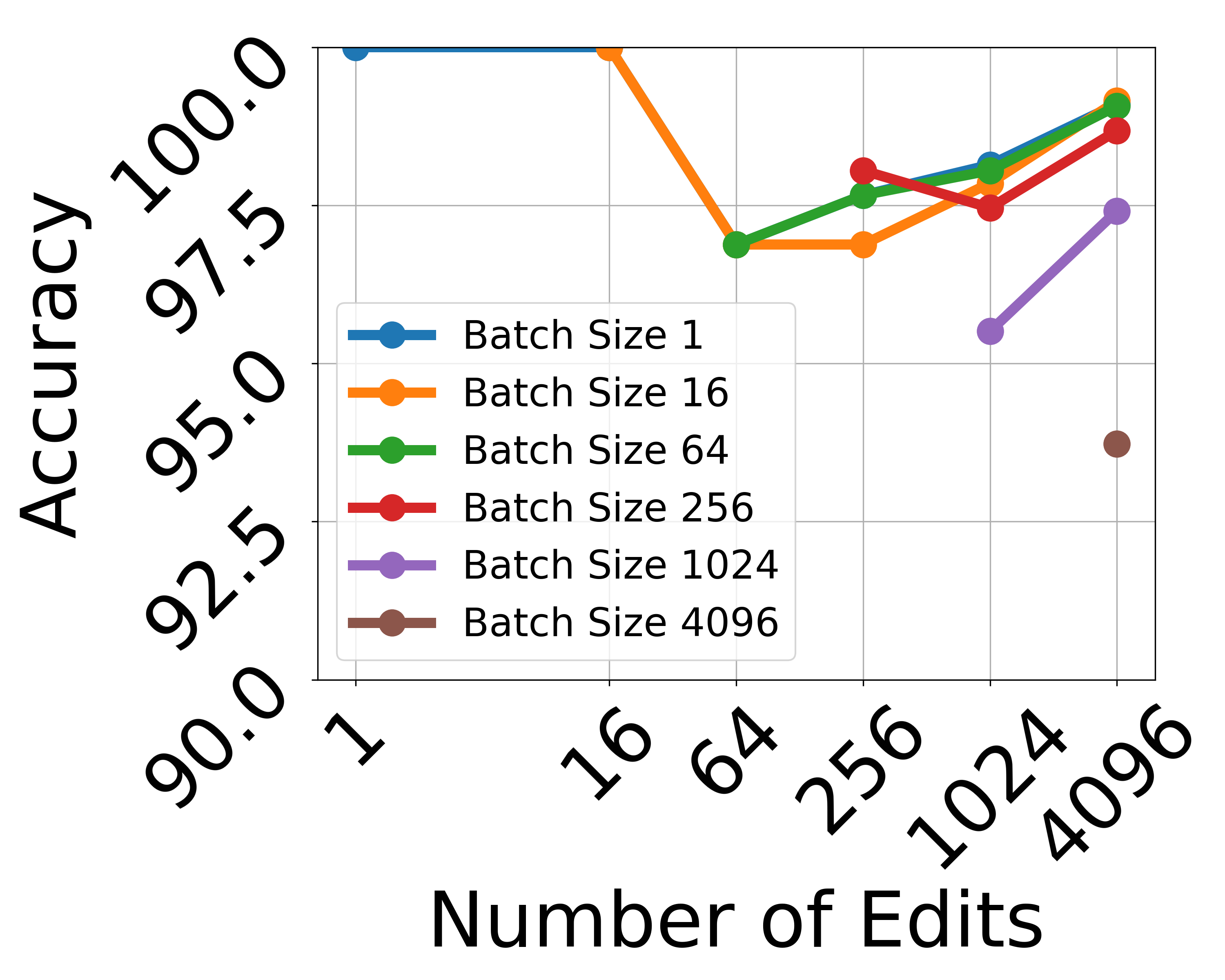}
        \caption{Efficacy Score (ES)}
        \label{fig:seqMEMITa}
    \end{subfigure}
    \begin{subfigure}[b]{0.24\textwidth}
        \centering
        \includegraphics[width=\textwidth]{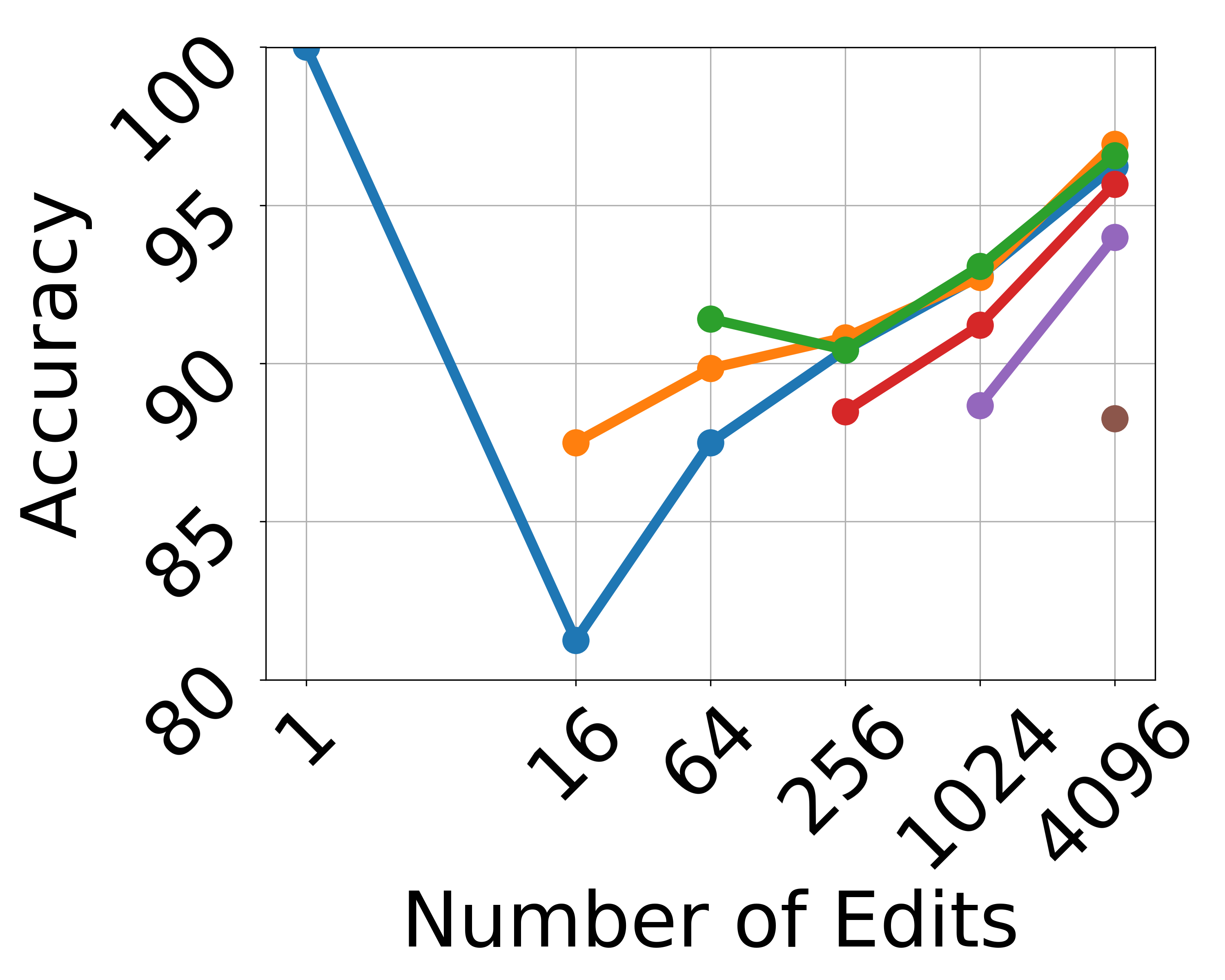}
        \caption{Paraphrase Score (PS)}
        \label{fig:seqMEMITb}
    \end{subfigure}
    \begin{subfigure}[b]{0.24\textwidth}
        \centering
        \includegraphics[width=\textwidth]{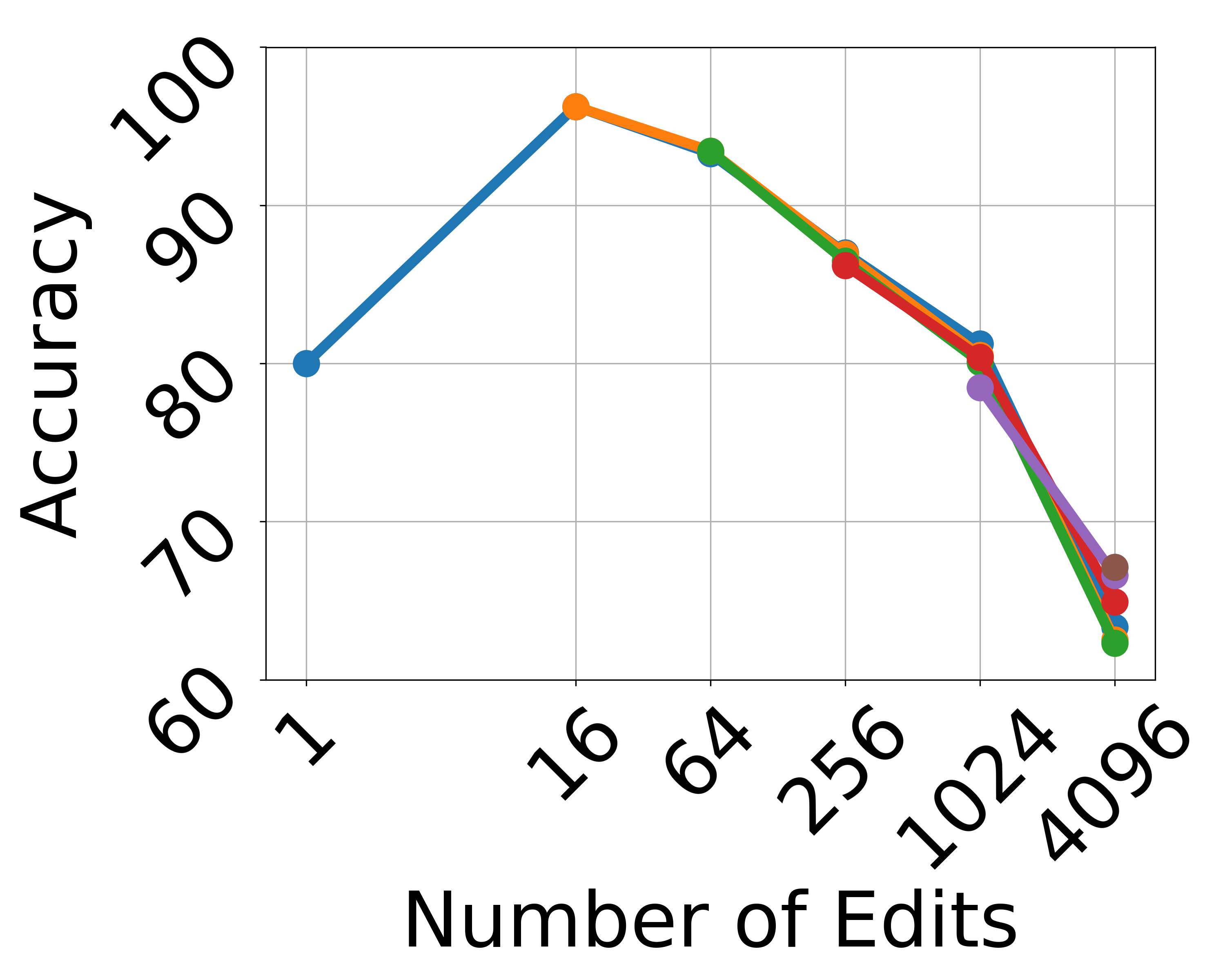}
        \caption{Neighborhood Score (NS)}
        \label{fig:seqMEMITc}
    \end{subfigure}
    \begin{subfigure}[b]{0.24\textwidth}
        \centering
        \includegraphics[width=\textwidth]{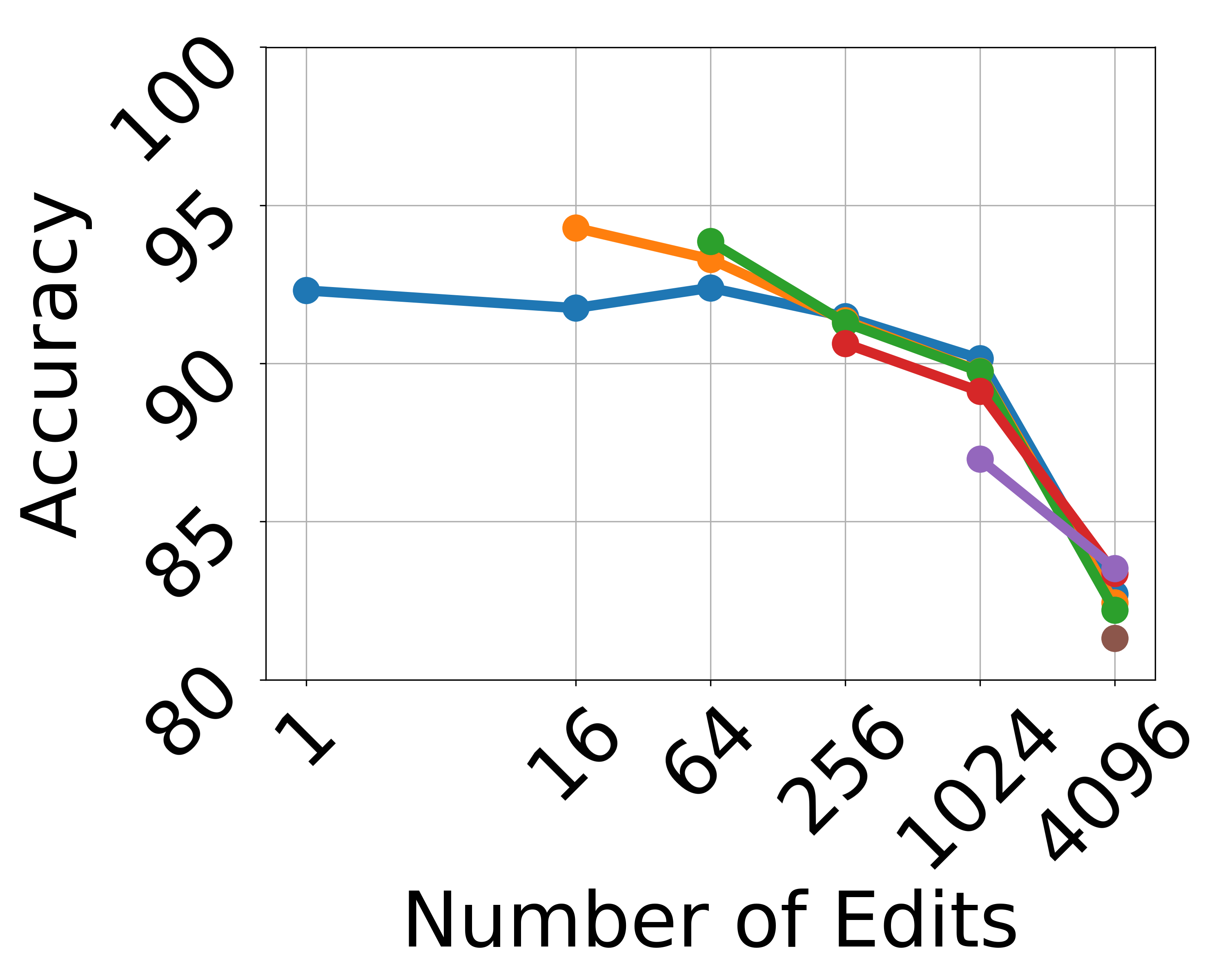}
        \caption{Score (S)}
        \label{fig:seqMEMITe}
    \end{subfigure}
    
    \caption{Single layer sequential editing performance of MEMIT for various batch sizes.}
    \label{fig:seqMEMIT}
\end{figure*}

\begin{figure*}
    \centering
    
    \begin{subfigure}[b]{0.24\textwidth}
        \centering
        \includegraphics[width=\textwidth]{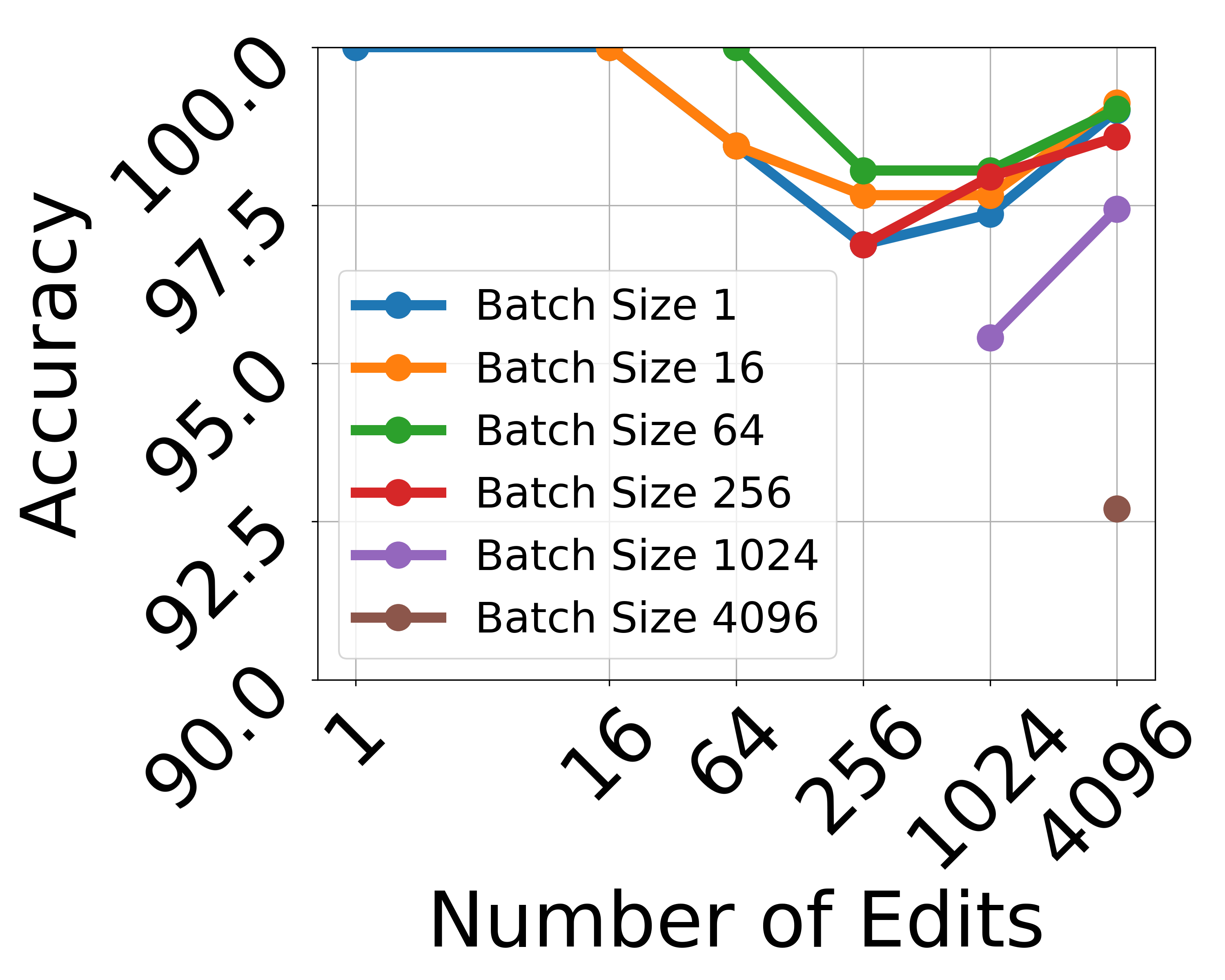}
        \caption{Efficacy Score (ES)}
        \label{fig:seqEMMETa}
    \end{subfigure}
    \begin{subfigure}[b]{0.24\textwidth}
        \centering
        \includegraphics[width=\textwidth]{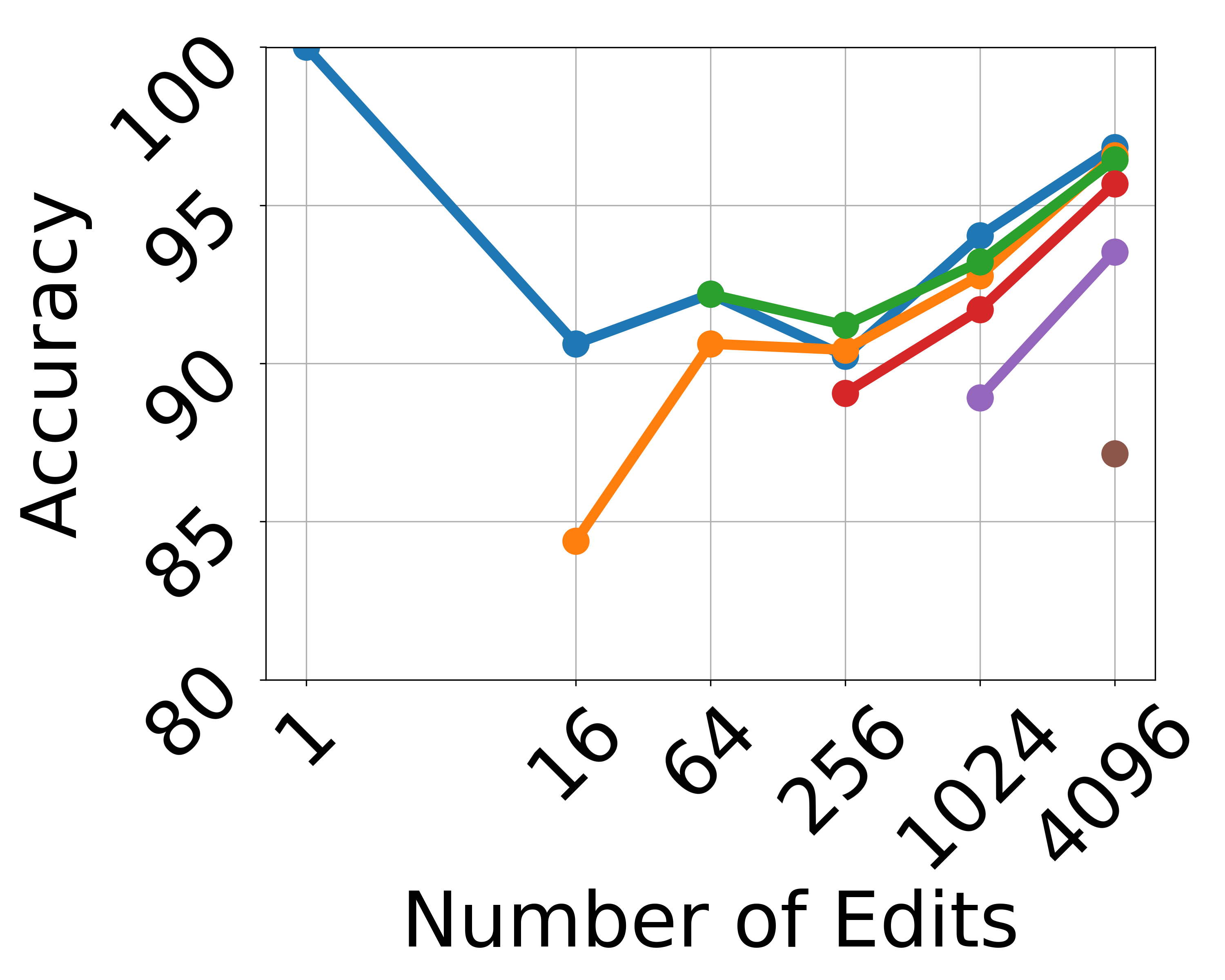}
        \caption{Paraphrase Score (PS)}
        \label{fig:seqEMMETb}
    \end{subfigure}
    \begin{subfigure}[b]{0.24\textwidth}
        \centering
        \includegraphics[width=\textwidth]{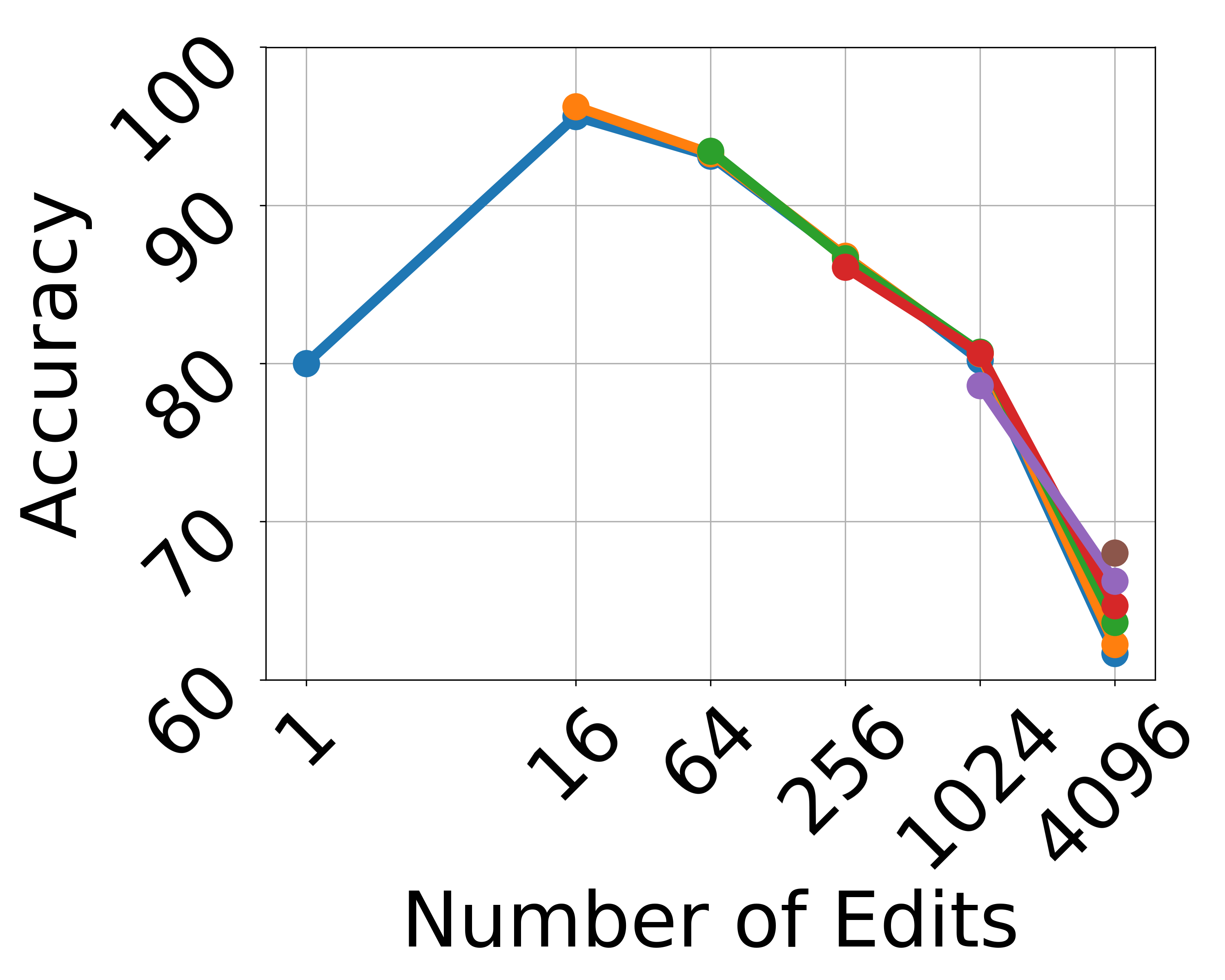}
        \caption{Neighborhood Score (NS)}
        \label{fig:seqEMMETc}
    \end{subfigure}
    \begin{subfigure}[b]{0.24\textwidth}
        \centering
        \includegraphics[width=\textwidth]{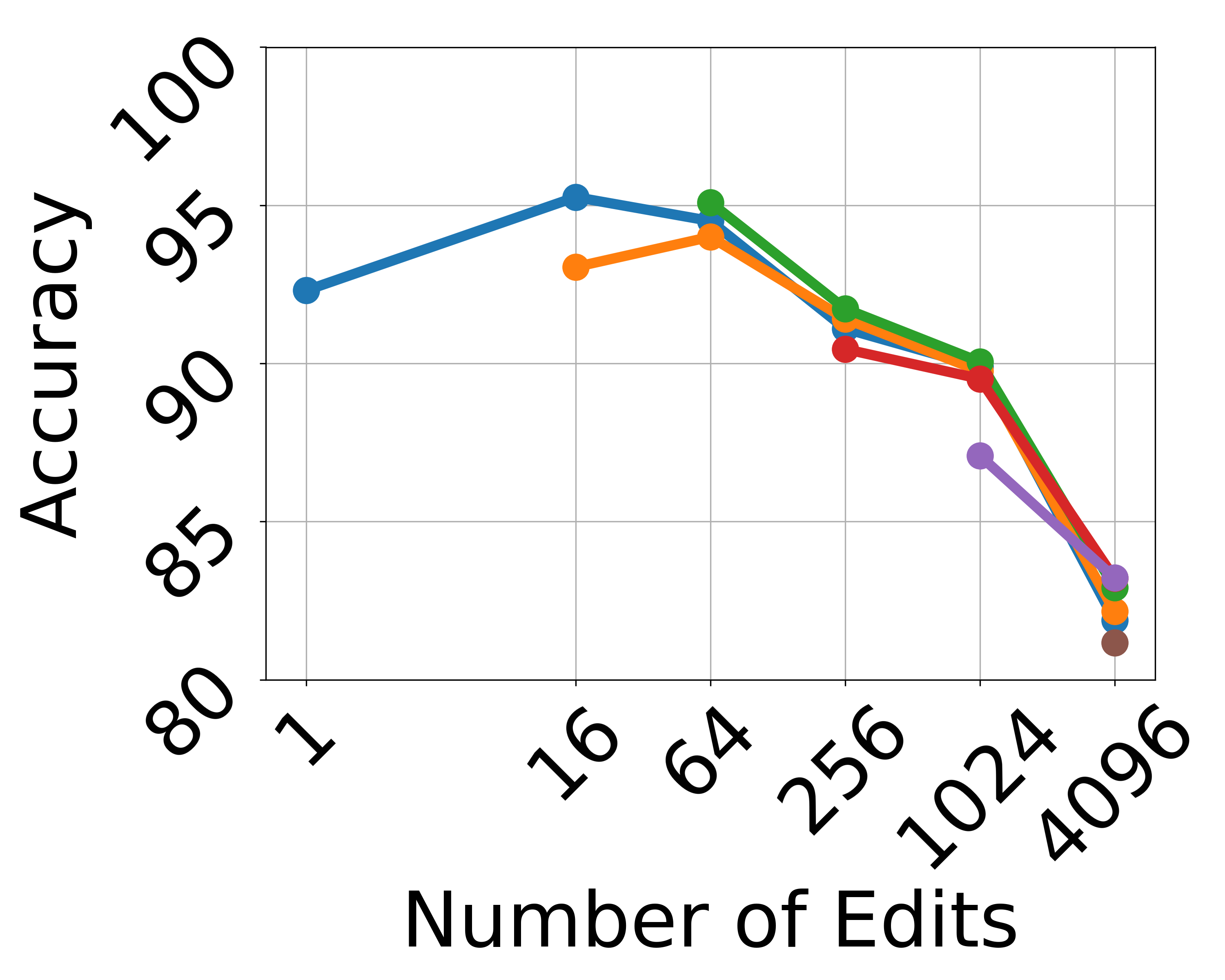}
        \caption{Score (S)}
        \label{fig:seqEMMETd}
    \end{subfigure}
    
    \caption{Single layer sequential editing performance of EMMET for various batch sizes}
    \label{fig:seqEMMET}
\end{figure*}

Prior works have studied sequential editing with batch size of 1, which means only one fact is updated with each gradient update \citep{editing-survey, akshat-catastrophic}. We generalize this idea to sequential-batched editing, where we update a batch of facts with one update, and sequentially edit many batches at a time, going from batch size of 1 up to 4096. We perform sequential-batched edits with varying batch sizes (1, 64, 256, 1024, 4096) using the MEMIT and EMMET editing methods, respectively, where batch size of 1 represents purely sequential edits. Figure \ref{fig:seqMEMIT} presents the outcomes of various metrics applied to the MEMIT method, while Figure \ref{fig:seqEMMET} examines the same for EMMET. Note that sequential batched edit with batch size of 1 corresponding to performing sequential editing with ROME. This comparative analysis aims to determine the most effective editing strategy for enhancing model accuracy and efficiency.


Figure \ref{fig:seqMEMITa}, \ref{fig:seqMEMITb}, \ref{fig:seqMEMITe} (Sequential MEMIT: ES, PS, S) suggests that \textbf{larger batch sizes are actually worse for model performance than sequential edits with smaller batches}. This can be seen by how one batch of 4096 edit is the worst performing batch size, when compared to sequential edits of smaller sizes. In contrast, larger batch sizes seem to be better for metrics in figure~\ref{fig:seqMEMITc} (Sequential MEMIT: NS). In this instance, the results seem to suggest that while batch edits are less successful in general, it is better in preserving locality of edits, where adjacent stored facts are not as strongly impacted. The same can be seen observations can be made for EMMET, as shown in Figure \ref{fig:batchEMMET}. Above plots also show an optimal batch size of 1024 for both MEMIT and EMMET. Increasing batch-size beyond that diminishes returns as larger batch sizes lead to larger model degradation and better editing results can be achieved by sequential-batched editing with smaller batch sizes.


\section{Conclusion}
Our study examines several model editing techniques in the context of the newly released Llama-3 model. Contrary to previous belief, our experiments show that earlier layers may be more optimal intervention points, and that smaller, frequent sequential batch size edits have a superior performance in comparison to larger batch sizes. Future work will include experiments on multi-layer intervention for edits, as well as experiments against other popular models and algorithms, including methods that are hyper-network based.



\bibliography{custom}
\newpage

\appendix

\section{Appendix}

\begin{table}[h]
    \centering
    \begin{tabular}{c|c|c} 

    \textbf{Batch Size} & \textbf{Num Batches} & \textbf{Total Edits} \\ \hline
    1 & 4096 & 4096 \\ \hline
    16 & 256 & 4096 \\ \hline
    64 & 64 & 4096 \\ \hline
    256 & 16 & 4096 \\ \hline
    1024 & 4 & 4096 \\ \hline
    4096 & 1 & 4096 \\ \hline
    \end{tabular}
    \caption{Statistics for batch size and number of batches used to create the numbers for this paper.}
    \label{tab:batch-size}
\end{table}

\subsection{Hyperparameter tuning}\label{sec:hparam-tuning}
Hyperparameter tuning was performed with batch size 1024 as the baseline for both MEMIT and EMMET algorithms, varying the update constant as shown in the background section. Table \ref{tab:hyperMEMIT} shows the hyperparameter sweep for MEMIT algorithm, from which lamba = 5e-4 was ultimately chosen, considering various metrics. Similarly, from Table \ref{tab:hyperEMMET},  lambda = 0 was chosen for the EMMET algorithm. 

\label{sec:appendix}

\begin{table*}[ht]
\centering
\caption{Hyperparameter Tuning for MEMIT Algorithm}
\label{tab:hyperparam_tuning_extended}
\begin{tabular}{@{}cccccc@{}}
\toprule
Lambda & ES & PS & NS & GE & S \\ \midrule
\(1\) & \(100.0 (0.0)\) & \(98.05 (11.34)\) & \(85.61 (21.93)\) & \(615.09 (32.01)\) & \(94.10\) \\
\(1 \times 10^1\) & \(100.0 (0.0)\) & \(98.05 (10.89)\) & \(85.73 (21.87)\) & \(614.42 (36.04)\) & \(94.15\) \\
\(1 \times 10^2\) & \(92.63 (26.13)\) & \(89.79 (24.72)\) & \(53.03 (32.23)\) & \(541.56 (70.60)\) & \(73.55\) \\
\(1 \times 10^3\) & \(93.36 (24.9)\) & \(89.43 (24.49)\) & \(52.86 (32.41)\) & \(521.97 (84.59)\) & \(73.51\) \\
\(1 \times 10^4\) & \(93.51 (24.64)\) & \(88.99 (25.14)\) & \(53.00 (32.30)\) & \(537.32 (72.02)\) & \(73.53\) \\
\(5 \times 10^4\) & \(97.71 (14.97)\) & \(92.7 (22.73)\) & \(73.19 (29.41)\) & \(611.48 (23.34)\) & \(86.49\) \\
\(1 \times 10^5\) & \(88.04 (32.45)\) & \(81.74 (34.71)\) & \(80.15 (25.77)\) & \(615.28 (21.18)\) & \(83.17\) \\
\(5 \times 10^5\) & \(43.6 (49.59)\) & \(39.79 (44.07)\) & \(87.05 (20.45)\) & \(616.78 (18.12)\) & \(50.37\) \\
\(1 \times 10^6\) & \(33.15 (47.08)\) & \(31.18 (41.44)\) & \(87.64 (20.04)\) & \(617.04 (17.98)\) & \(40.73\) \\
\bottomrule
\end{tabular}
    \label{tab:hyperMEMIT}

\end{table*}
\label{sec:appendix}

\begin{table*}
\centering
\caption{Hyperparameter Tuning for EMMET Algorithm }\label{tab:hyperEMMET}
\begin{tabular}{@{}cccccc@{}}
\toprule
Lambda & ES  & PS  & NS & GE & S \\ \midrule
\(0\) & \(94.09 (23.58)\) & \(90.14 (24.06)\) & \(53.5 (31.31)\) & \(545.84 (68.7)\) & \(74.23\) \\
\(1 \times 10^{-2}\) & \(70.56 (45.58)\) & \(68.16 (38.93)\) & \(52.56 (35.89)\) & \(563.84 (58.55)\) & \(62.67\) \\
\(1 \times 10^{-3}\) & \(79.39 (40.45)\) & \(76.27 (34.24)\) & \(51.39 (33.96)\) & \(487.69 (93.56)\) & \(66.42\) \\
\(1 \times 10^{-4}\) & \(89.99 (30.01)\) & \(85.72 (27.64)\) & \(52.08 (31.7)\) & \(522.15 (77.4)\) & \(71.46\) \\
\(1 \times 10^{-5}\) & \(92.33 (26.61)\) & \(86.72 (27.19)\) & \(51.86 (31.14)\) & \(530.25 (75.91)\) & \(72.04\) \\
\(1 \times 10^{-6}\) & \(93.60 (24.47)\) & \(90.75 (23.46)\) & \(52.08 (31.81)\) & \(529.64 (80.43)\) & \(73.34\) \\
\(1 \times 10^{-7}\) & \(93.95 (23.85)\) & \(90.11 (24.13)\) & \(52.05 (31.42)\) & \(529.02 (82.7)\) & \(73.25\) \\
\bottomrule
\end{tabular}
\end{table*}

\label{sec:appendix}

\subsection{Llama-2 Layer Search}

\begin{figure*}[!ht]
    
    \centering
    \begin{subfigure}{.24\textwidth}
        \centering
        \includegraphics[width=\linewidth]{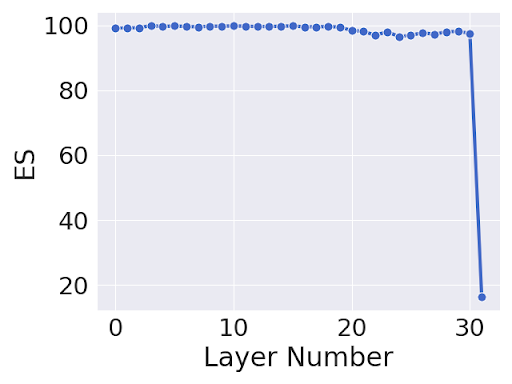}
        \caption{MEMIT-Llama2 ES}
    \end{subfigure}%
    \begin{subfigure}{.24\textwidth}
        \centering
        \includegraphics[width=\linewidth]{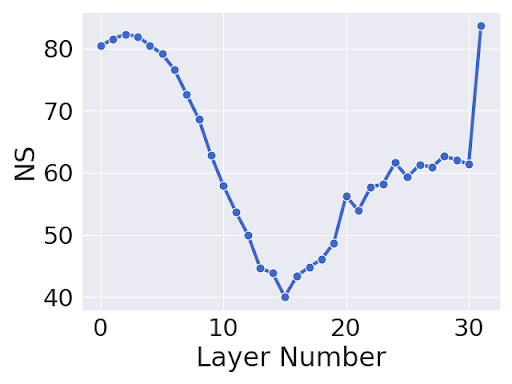}
        \caption{MEMIT-Llama2 NS}
    \end{subfigure}%
        \begin{subfigure}{.24\textwidth}
        \centering
        \includegraphics[width=\linewidth]{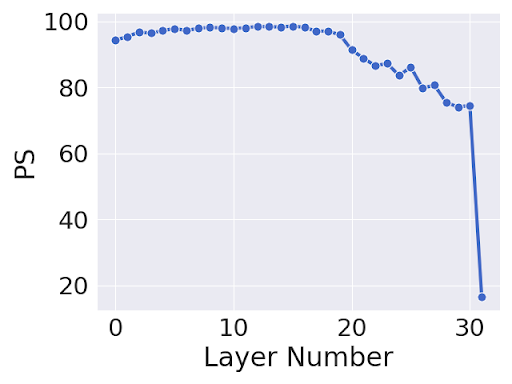}
        \caption{MEMIT-Llama2 PS}
    \end{subfigure}%
    \begin{subfigure}{.24\textwidth}
        \centering
        \includegraphics[width=\linewidth]{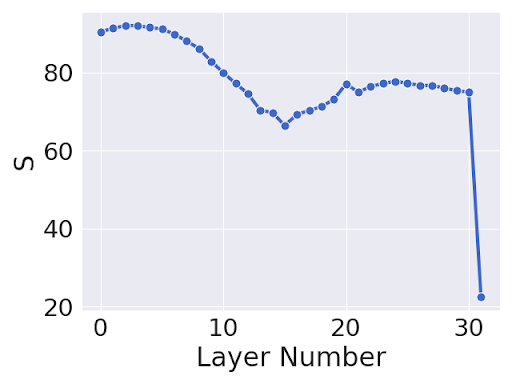}
        \caption{MEMIT-Llama2 S}
    \end{subfigure}

    \caption{Post-edit performance of various metrics on Llama2-7b for MEMIT on various layers. }
    \label{fig:layerwise-llama2}
\end{figure*}

Results for Llama-2 layer performance is shown in figure \ref{fig:layerwise-llama2}. 

\end{document}